# Novelty-Driven Target-Space Discovery in Automated Electron and Scanning Probe Microscopy

Utkarsh Pratiush[1*], Kamyar Barakati[1], Boris N. Slautin[1],

Catherine C. Bodinger,[2] Christopher D. Lowe,[2] Brandi M. Cossairt[2], Sergei V. Kalinin[1,3*]

[2] *Department of Materials Science and Engineering, University of Tennessee, Knoxville, TN 37996, USA*

[2] *Department of Chemistry, University of Washington, Seattle, WA 98195, USA*
[3] *Pacific Northwest National Laboratory, Richland, WA 99354, USA*
* Corresponding author

Modern automated microscopy faces a fundamental discovery challenge: in many systems, the most important scientific information does not reside in the immediately visible image features, but in the target space of sequentially acquired spectra or functional responses, making it essential to develop strategies that can actively search for new behaviors rather than simply optimize known objectives. Here, we developed a deep-kernel-learning BEACON framework that is explicitly designed to guide discovery in the target space by learning structure–property relationships during the experiment and using that evolving model to seek diverse response regimes. We first established the method through demonstration workflows built on pre-acquired ground-truth datasets, which enabled direct benchmarking against classical acquisition strategies and allowed us to define a set of monitoring functions for comparing exploration quality, target-space coverage, and surrogate-model behavior in a transparent and reproducible manner. This benchmarking framework provides a practical basis for evaluating discovery-driven algorithms, not just optimization performance. We then operationalized and deployed the workflow on STEM, showing that the approach can transition from offline validation to real experimental implementation. To support adoption and extension by the broader community, the associated notebooks are available, allowing users to reproduce the workflows, test the benchmarks, and adapt the method to their own instruments and datasets.

# I. Introduction

Electron microscopy[1–3] has emerged as a foundational tool for materials discovery, by providing direct access to structure and functionality at the length scales where materials behavior ultimately emerges. In recent years, rapid advances in measurement modalities[4,5] ranging from Electron Energy Loss Spectroscopy (EELS)[6], off-axis EELS, and time-delay cathodoluminescence have substantially expanded what can be learned from a single instrument platform. Collectively, these developments enable a growing set of contrast mechanisms providing insight into inner-shell electronic transitions, orbital occupancies, phonon, and band structure that were previously associated with macroscopic characterization, but are now accessible at the atomic scale[7–11]. In effect, many electron microscopy modalities are transferring established characterization concepts into regimes where individual atomic environments can be interrogated with unprecedented specificity. This convergence is opening fundamentally new opportunities[12–14] to explore matter at the atomic level, not only by visualizing structure, but by probing local excitations and responses with increasing precision[15].

To date, many of the most advanced STEM capabilities have been demonstrated first on idealized, well-defined objects such as interfaces, topological defects, and other features whose structure and location can be anticipated in advance[16–20]. These systems provide clear, interpretable targets for method development and validation[21–23]. However, real materials are typically characterized by complex, heterogeneous microstructures in which the objects of interest are not established a priori[24,25], and may not even be readily recognizable before measurement. In this setting, the central challenge shifts from measuring a known feature to discovering which features matter and where they reside within a large and intricate atomic- or mesoscopic landscape. At the same time, practical constraints[22] impose hard limits on brute-force strategies. The time required for acquisition, combined with beam damage[26,27] considerations, preclude measurements over dense grids followed by retrospective analysis, even if such exhaustive sampling would be conceptually attractive.

These considerations necessitate the development of machine learning algorithms that can be deployed on automated microscopes to identify and prioritize regions of interest as the experiment proceeds[28–32]. Rather than relying on dense, uniform sampling followed by offline interpretation, such algorithms are intended to guide measurement toward the most informative locations within complex microstructures, where the relevant objects are not known in advance.

Over the last several years, parallel progress in machine learning methods and in instrument-control APIs[33–41] has made this direction technically feasible, enabling the integration of ML-driven decision logic with microscope automation and thereby supporting the practical implementation of automated region selection in STEM experiments.

In general, automated discovery of structure–property relationships begins[22,31,42,43] from the situation in which the structural data are available in full, and the locations for subsequent spectral measurements are then selected sequentially based on the structural information and the spectra acquired thus far. Within this broad framing, a large class of strategies operates on the structural data alone: measurement locations are chosen based on *a priori* known objects of interest, statistical reweighting of the observed structures, or some combination of novelty scores defined over the structural observations. In these scenarios[24,30,44–46], navigation of the feature space, typically represented as image patches, is effectively decoupled from the measured spectra: the trajectory through structural variations is not directly altered by what is learned from the spectral measurements themselves.

An alternative set of approaches is built around discovery in the spectral domain. Here, the key distinction is that spectral data are available only sequentially, so the methodology must explicitly account for the evolving nature of what is known about the structure–property relationship as measurements accumulate. These algorithms also require selection of the discovery target, typically a scalar functional (scalarizer)[43] of measured spectra via established physical knowledge[22] to provide a clear discovery target.[43,45,60–62] The core of these methods is therefore active learning of a surrogate model connecting local structure to spectra or scalarizer that is dynamically updated during the experiment, and whose probabilistic predictions are used to select the next measurement points. Typical examples include variants of deep kernel learning methods that directly construct such surrogate models[42,47–50], as well as combinations of dimensionality reduction and regression, where the dimensionality-reduction stage can include linear methods such as principal component analysis (PCA)[51] and non-linear dimensionality reduction via variational autoencoders (VAE)[52,53], and the regression stage can employ Gaussian processes[54], ensembled networks, or other models[55] that yield prediction and uncertainty required to build an acquisition function for Bayesian optimization (BO)[56].

Both the feature-space and spectral-space methods also accommodate human-in-the-loop intervention[43,57–59]. In practice, a human operator can influence feature-space strategies by prioritizing the selection of particular objects in the feature space, and can influence spectral-space strategies by refining the scalarizer functions used to define targets in the spectral domain. In addition, in both cases the operator can tune[59] the exploration–exploitation balance, shaping how aggressively the system pursues novelty versus consolidates around regions expected to yield high-value measurements.

However, to date, one of the missing constraints has been the design of the discovery function itself. Classical Bayesian optimization models require a combination of exploration, typically defined through the uncertainty of prediction, and exploitation defined through a known objective function, in order to construct an acquisition function. In this framing, the uncertainty of prediction is defined over the feature space. In many cases, of interest is not the discovery of regions with specific scalarizer values or minimization of the prediction uncertainty in the feature space, but discovery of the possible behaviors in the target space irrespectively of how strongly these behaviors are present, i.e. the algorithms that allow to explore novelty[48–50] directly in the target space[63–65]. Here we demonstrate the extension of the BEACON algorithm[63] for novelty discovery implemented on STEM in the deep kernel learning model[43,55].

## II. BEACON DKL: Framework for Autonomous Discovery

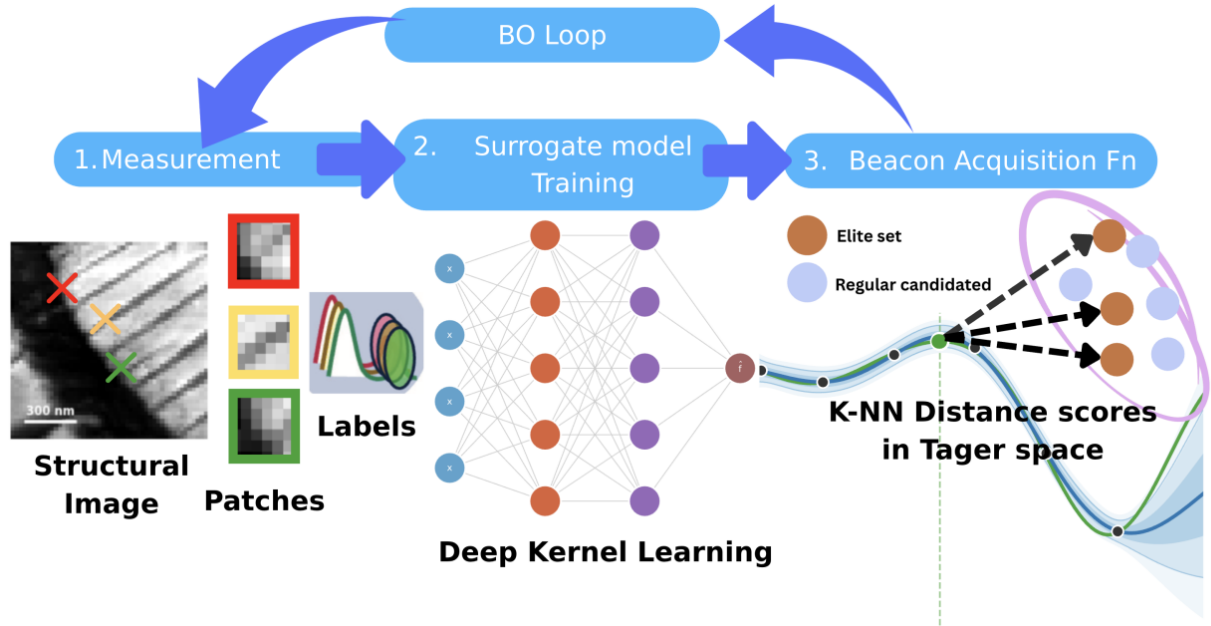

**Figure 1.** Schematic of the BEACON active learning loop. At each iteration, image patches and their scalarized labels are used to train a deep kernel learning surrogate model. The BEACON acquisition function selects the next measurement using the nearest neighbor-based novelty score. The loop repeats until the measurement budget is exhausted.

The transition from identifying known structures to uncovering emergent physical phenomena requires a computational framework that can navigate the vast, often non-linear relationship between atomic configuration and functional response. In this section, we detail the implementation of a discovery-driven autonomous loop that integrates the high-dimensional feature extraction capabilities of Deep Kernel Learning (DKL) with the BEACON (Bayesian Evolutionary Analysis for Cosmological Observation Networks) acquisition protocol. By moving beyond traditional value-maximizing and uncertainty-based strategies, this framework utilizes Thompson Sampling to stochastically explore the latent target space, prioritizing regions of structural interest not by their proximity to a predefined goal, but by their potential to yield statistically novel physical signatures as reflected in the spectrum of possible scalarizer values. This approach allows to explore the novelty in the feature space of the system, as well as partially alleviates the "scalarizer bottleneck", the requirement for deep domain expertise to define a reward function a priori, enabling the microscope to autonomously identify rare defects and emergent phases in complex, heterogeneous microstructures.

While the original BEACON framework[63] provides a generalized mathematical approach for searching for novelty in black-box systems, our work (schematic in **Figure 1)** adapts this logic into a specialized Deep Kernel Learning (DKL) architecture designed for high-dimensional microscopy data. The primary contribution lies in replacing standard parameter-based inputs with a CNN-driven feature extractor that enables the discovery engine to process raw HAADF-STEM image patches directly. The method involves computing a novelty score by implementing a stochastic k-nearest neighbor (k-NN) distance metric measured against a dynamically updated elite set, specifically tuned for the "unknown unknowns" of atomic-scale physics. We note that here we implement the novelty discovery in the scalarizer space. In principle, the novelty can be also defined using the high-dimensional feature space of the measured spectra, but in this case novelty definition becomes dependent on the chosen high-D metric function (distance between spectra), making the discovery less explicit then scalarizer based definition.

## II.a. Deep Kernel Learning (DKL) Architecture

To bridge the gap between raw real-space imaging and spectral prediction, we employ a DKL framework. A Convolutional Neural Network (CNN), acting as a feature extractor $\phi(x)$, compresses high-dimensional structural patches (e.g., $16 \times 16$ pixel sub-images) into a low-dimensional latent space. This latent representation serves as the input for a Gaussian Process (GP). The kernel function $k(x, x')$ is transformed into a deep kernel:

$$k_{\text{deep}}(x, x') = k(\phi(x, w), \phi(x', w)) \qquad (1)$$

where $w$ represents the weights of the CNN. By training the CNN and GP hyperparameters simultaneously through the maximization of the marginal log-likelihood, the model learns a structural embedding optimized to predict the local physical response.

## II.b. The BEACON Acquisition Function

The core of our discovery engine is the BEACON acquisition function as described in Equation 2. Unlike standard protocols that maximize a scalar value, BEACON seeks novelty by measuring distances in the target (response) space. For a set of acquired points $\mathcal{D}_n$, we define an "elite set" $\mathcal{E} \subset \mathcal{D}_n$ containing the top fraction of measurements based on a user-defined physical criteria. For a candidate location $x$, the acquisition value $\alpha_{\text{BEACON}}(x)$ is defined as the average distance to its $k$-nearest neighbors in the elite set:

$$\alpha_{\text{BEACON}}(x) = \frac{1}{k}\sum_{j=1}^{k}|\hat{y}(x) - y_j|, \quad y_j \in \text{NN}_k(\hat{y}(x), \mathcal{E}) \qquad (2)$$

where $\hat{y}(x)$, is a sample from the posterior distribution at $x$, and $y_j$ are the responses of the $k$ most similar points in the elite set.

### II.c. Stochastic Exploration via Thompson Sampling

To robustly handle model uncertainty, we utilize Thompson Sampling (TS). At each step, instead of using the posterior mean $\mu(x)$, we draw a sample $\hat{y}(x)$ from the full posterior distribution:

$$\hat{y}(x) \sim \mathcal{N}(\mu(x), \sigma^2(x)) \qquad (3)$$

By evaluating novelty on these samples, the system naturally balances exploration and exploitation. If a region has high uncertainty ($\sigma^2$), the samples $\hat{y}$ will vary widely, occasionally producing "novel" values that trigger the microscope to explore that structural regime.

### II.d. The BEACON-DKL Algorithm

The following algorithm outlines the autonomous discovery loop implemented in our study.

```
Algorithm 1: BEACON-DKL
Input  : Search space $\mathcal{X}$, Initial seed budget $N_{\text{seed}}$, Total budget
Output : Final dataset $\mathcal{D} = \{(\mathbf{x}_i, y_i)\}$
Initialize: Acquire $N_{\text{seed}}$ random points to form $\mathcal{D}$. Initialize NN $\phi_{\mathbf{w}}$
           and GP kernel $k_\theta$
while experimental budget remains do
    1. Joint Training:
       Minimize Negative Log-Likelihood: $\mathcal{L}(\mathbf{w}, \theta) = -\log p(\mathbf{y}|\mathbf{X}, \mathbf{w}, \theta)$
       (Updates NN weights $\mathbf{w}$ and GP hypers $\theta$ simultaneously)
    2. Feature Extraction (Latent Mapping):
       Map training inputs: $\mathbf{Z}_\mathcal{D} = \{\phi_{\mathbf{w}}(x) \mid x \in \mathcal{D}\}$
       Map search space: $\mathbf{Z}_\mathcal{X} = \{\phi_{\mathbf{w}}(x) \mid x \in \mathcal{X}_{\text{unq}}\}$
    3. DKL Posterior Inference:
       Construct Covariance Matrix: $K_{\mathbf{ZZ}} = k_\theta(\mathbf{Z}_\mathcal{D}, \mathbf{Z}_\mathcal{D}) + \sigma_n^2 I$
       for each candidate feature $\mathbf{z} \in \mathbf{Z}_\mathcal{X}$ do
           $\mu(\mathbf{z}) = k_\theta(\mathbf{z}, \mathbf{Z}_\mathcal{D}) K_{\mathbf{ZZ}}^{-1} \mathbf{y}$
           $\sigma^2(\mathbf{z}) = k_\theta(\mathbf{z}, \mathbf{z}) - k_\theta(\mathbf{z}, \mathbf{Z}_\mathcal{D}) K_{\mathbf{ZZ}}^{-1} k_\theta(\mathbf{Z}_\mathcal{D}, \mathbf{z})$
       end
    4. Thompson Sampling:
       Draw posterior sample $\hat{y}(x) \sim \mathcal{N}(\mu(\phi_{\mathbf{w}}(x)), \sigma^2(\phi_{\mathbf{w}}(x)))$
       Normalize sample using elite set $\mathcal{E}$ statistics: $\hat{y}_{\text{norm}}$
    5. Rank & Acquire:
       Compute novelty $\alpha_{\text{BEACON}}$ as $k$-NN distance in latent space $\mathbf{Z}$
       Identify $x^* = \arg\max \alpha_{\text{BEACON}}(x)$
       Execute measurement $y^*$ at $x^*$
    6. Update:
       $\mathcal{D} \leftarrow \mathcal{D} \cup \{x^*, y^*\}$
end
return $\mathcal{D}$
```

As illustrated in **Algorithm 1**, the discovery process begins by initializing a search space with a small set of random seed points to establish a baseline dataset. At each iteration of the experimental loop, a Deep Kernel Learning (DKL) model is trained on the currently acquired data, optimizing the convolutional neural network's feature extraction and the Gaussian Process hyperparameters simultaneously. Once trained, the algorithm identifies an elite set by ranking the acquired points according to their physical response and selecting the top-performing fraction as a reference for known "interesting" physics. This elite set is then used to compute z-score normalization parameters, the mean and standard deviation, which define the statistical bounds of established high-value behavior.

To select the next measurement point, the model generates predictions for all unacquired structural patches; however, rather than relying on a deterministic mean, it employs Thompson Sampling to draw stochastic realizations from the full posterior distribution. These samples are normalized using the elite set's statistics and then compared against the elite distribution using a $k$-nearest neighbor ($k$-NN) distance calculation in the target space. The resulting BEACON novelty score prioritizes candidates that are predicted to yield physical responses furthest from the current elite cluster. The microscope then autonomously acquires a spectrum at the location with the highest novelty score, updates the dataset with this new structural-property pair, and repeats the cycle until the experimental budget is exhausted.

## II.e. Discussion on Latent Mapping

The power of this approach lies in the latent mapping of the target space. In traditional BO, the "distance" is measured in the input space (structural similarity). However, in materials science, two structures that look different might produce identical physics, or subtle structural changes might trigger phase transitions with massive spectral shifts. By using the Mahalanobis-like distance in the spectral response space, BEACON effectively decorrelates the discovery process from structural novelty. It focuses the experimental budget strictly on regions where the predicted physical behavior is statistically "extreme" compared to the current elite distribution. This allows for the discovery of rare defects or emergent phases that would be overlooked by mean-seeking algorithms.

## II.f. Implementation details

The active learning framework is implemented in Python, leveraging GPyTorch[66] for the deep kernel learning (DKL) surrogate model and BoTorch[56] for the Bayesian optimization loop. The surrogate model combines a convolutional neural network (ConvNet) feature extractor with a stochastic variational Gaussian process (SVGP), enabling structure-aware uncertainty quantification directly from image patches. Candidate point selection is performed via a custom BEACON acquisition function inspired from[63], which uses Thompson sampling and k-nearest-neighbor distances in the posterior space to promote novelty-driven exploration of the sample. Prior to deployment on a physical instrument, the full active learning pipeline is validated against preacquired datasets using a digital twin microscope[67]. Microscope hardware control is handled through the AutoScript TEM Microscope Client API, wrapped within the stemOrchestrator library[33,34], which manages HAADF image acquisition, beam positioning, and EDS spectrum collection across four detector channels. At each active learning step, the beam is directed to the selected spatial coordinate, an EDS spectrum is acquired with a fixed live-time exposure, and the summed spectral counts (or element-specific peak intensities) serve as the scalar objective fed back to the BO loop. The workflow is hardware-agnostic, running on either CPU or GPU, though GPU acceleration is recommended for practical experimental throughput.

**III. Validation on SPM and STEM**

To illustrate the novelty discovery via BEACON DKL, we first apply this approach to pre-acquired data sets for electron and scanning probe microscopy, introduce monitoring functions that define the rates of novelty discovery/optimization, and establish initial benchmarks. These workflows are also provided in notebooks accompanying the publication and can be used to deploy directly to instruments once the APIs/clients are available.

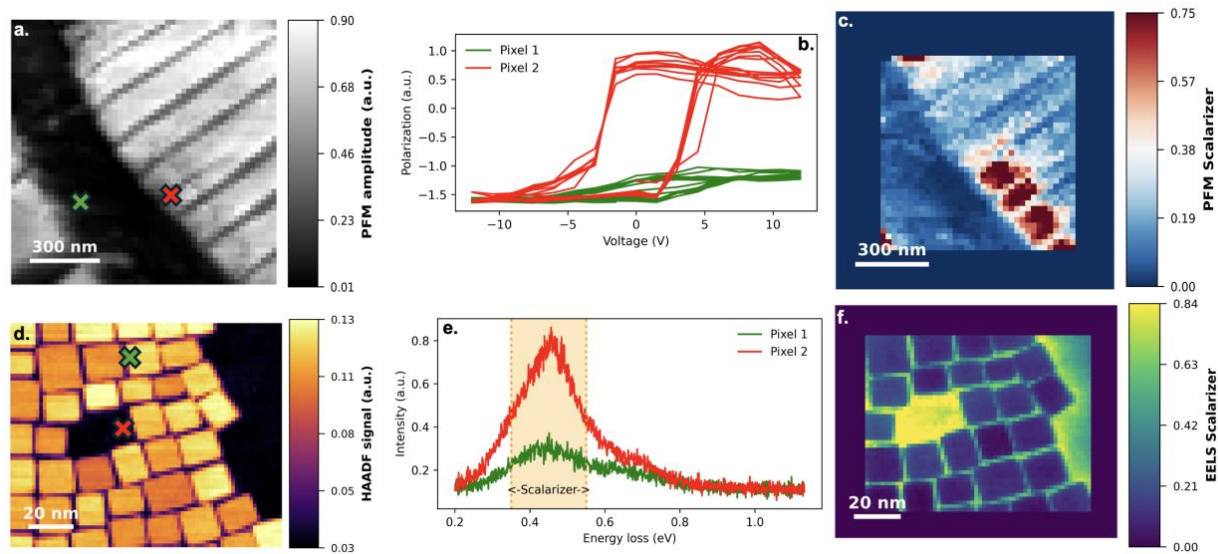

**Figure 2.** Example structure–property workflows used for active learning in scanning probe and electron microscopy. (a) Topographic PFM image of a ferroelectric PTO surface with two example pixel locations. (b) Spectroscopic PFM measurements by PFM provide the information of polarization switching mechanisms, (c) Ground-truth PFM scalarizer map derived from the spectroscopic response across the scan area. (d) HAADF-STEM image of plasmonic nanoparticles with two selected probe locations. (e) Corresponding EELS spectra from the selected pixels highlighting variations in plasmonic response within the scalarizer window. (f) Ground-truth scalarizer map derived from the EELS signal. These scalarizer fields represent the target properties that active learning algorithms aim to discover efficiently by sampling the feature space while minimizing the number of measurements.

To illustrate the principles and practical implementation of the BEACON algorithm, **Figure 2** presents two representative systems for which ground-truth data are available. The first example comes from scanning probe microscopy, specifically a piezoresponse force microscopy (PFM) image of ferroelectric domains in the lead titanate system, where the domain pattern is clearly resolved. This materials system has been broadly used previously to explore and operationalize the automated SPM workflows[31,46,68]. Here, dark regions correspond to the in-plane *a* domains, whereas bright regions correspond to out of plane *c* domains. The PFM contrast serves as the fast-to-acquire structural image. In the same experiment, the microscope can also run spectroscopic measurements by measuring local hysteresis loops. **Figure 2b** shows example loops acquired at several locations. Here, the discovery target can be posed directly in terms of how the hysteresis loop behavior varies across the sample.

Under special circumstances, these hysteresis loops can be collected on a dense grid (e.g., on the order of 80 × 80 measurements), providing a ground-truth map of the loop response. Because direct visualization of the full hysteresis-loop dataset is cumbersome, it is common to focus on a scalar representation. For example, **Figure 2c** shows a map of the hysteresis-loop area. In a classical grid-mode experiment, loops are acquired everywhere, and the resulting maps are analyzed to reveal correlations between local microstructural features and polarization switching behavior, including associations with dislocations. However, dense-grid acquisition is time-consuming and often carries a significant risk of probe and sample damage. The role of active learning, in this setting, is therefore to begin from the fully available PFM image and sequentially acquire hysteresis loops at new locations so as to learn the structure–property relationship of interest, in this case the relationship between local microstructure and hysteresis-loop behavior, without resorting to exhaustive sampling.

**Figure 2d-f** provide an analogous example for a STEM-EELS dataset. **Figure 2d** shows a dark-field image of plasmonic nanostructures[69], while the corresponding EELS spectra are illustrated alongside, including the definition of a scalarizer function. The latter is chosen to be dipole plasmon mode. **Figure 2e** then shows the ground-truth spatial behavior of this scalarized target. As in the PFM case, the central active-learning problem is set by the asymmetry between what is available and what is costly: the structural information is available everywhere, whereas spectral measurements are obtained sequentially. The decision of which point to measure next is therefore driven by the overall reward function balancing discovery and optimization, while leveraging the available structural data to extract maximal information from a limited number of spectra.

Historically, active learning in these systems has been explored primarily through the lens of optimization: classical decision functions are used to favor either exploration of the object space or exploitation/optimization in the feature space. In practice, however, these strategies often exhibit a pronounced tendency to collapse onto a localized region of the image space, effectively concentrating measurements in one area and prematurely arresting exploratory activity. To address this behavior, we previously introduced a human-in-the-loop approach[43,57] in which the reward function can be dynamically adjusted during the experiment, alongside several forms of novelty discovery algorithms as outlined above. Here, we illustrate exploratory behavior using the BEACON algorithm, which explicitly favors novelty discovery in the feature space—that is, it is designed to seek out regions that exhibit distinct behavior of the scalarizer function, rather than converging rapidly onto a single local optimum.

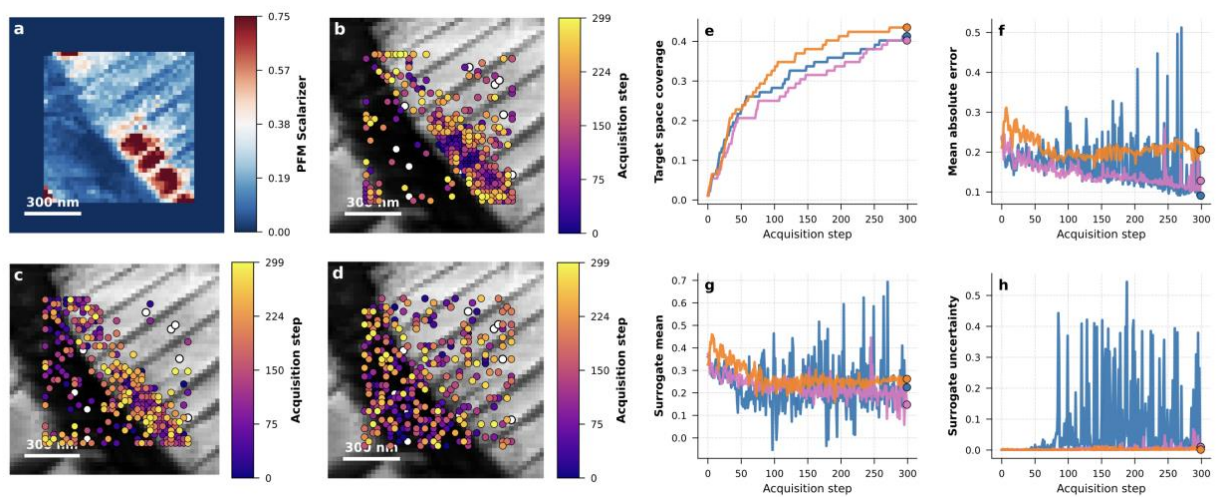

**Figure 3.** Comparison of active learning strategies on PFM data. (a) Ground-truth PFM scalarizer map. (b–d) Acquisition trajectories for EI (b), MU (c), and BEACON (d), with marker color encoding step order and white markers denoting seed points. (e) Target space coverage **(see definitions in supplementary)**, (f) surrogate MAE, (g) surrogate mean, and (h) surrogate uncertainty as a function of acquisition step. Colors: EI (blue), MU (pink), BEACON (orange).

To compare classical deep-kernel-learning-based active learning using the expected improvement (EI) and Maximum Uncertainty (MU) acquisition functions against BEACON, **Figure 3** shows the ground-truth reward landscape together with the corresponding exploration trajectories over 300 active-learning steps. For both EI and MU as shown in **Figure 3a-b,** there is a clear tendency for the trajectory to concentrate within a limited part of the image space, in this case associated with anomalously high values of the acquisition function. In practice, the first several steps sample the image relatively broadly, but this early exploratory phase is followed by a strong collapse of the trajectory onto a narrow spatial region. This effect is most pronounced for EI and remains visible, although less severe, for MU. An important visual signature of this behavior is that the colors encoding the time order of the trajectory form well-defined clusters for both EI and MU. Operationally, this means that once the algorithm reaches a favorable region, it begins to sample that region repeatedly rather than continuing to traverse the broader landscape.

This observation is particularly important because the algorithms do not use the explicit spatial coordinates of the patch as an input. Instead, they operate only on the relationship between the local microstructure, represented by the image patch, and the corresponding scalarized response. Therefore, concentration of the trajectory in a localized region of the $x$-$y$ image plane is not imposed by the model directly; rather, it emerges because either (i) the relevant behaviors are strongly concentrated in that part of the sample, or (ii) the acquisition policy has become effectively trapped in a restricted region of the learned feature space. During the experiment, these two possibilities cannot be cleanly separated. In either case, however, the practical consequence is the same: the exploratory process loses diversity and ceases to probe the sample broadly. In contrast, BEACON (**Figure 3c**) produces a much more spatially distributed trajectory, and the colors corresponding to acquisition time appear substantially more intermixed. This qualitative behavior indicates that the algorithm continues to interrogate different parts of the sample surface instead of repeatedly returning to a single narrow region.

Target space coverage (**Figure 3e**) measures the fraction of the ground-truth scalarizer distribution that has been visited as a function of acquisition step. BEACON reaches higher coverage earlier and maintains a consistent lead throughout the 300-step budget, indicating that its trajectories sample a broader range of physical responses rather than repeatedly probing the same narrow value range. EI and MU converge toward similar final coverage values but do so more slowly and with a shallower initial rise, consistent with the trajectory collapse observed in the spatial maps. The mean absolute error of the surrogate model (**Figure 3f**) evolves differently across the three strategies. EI exhibits pronounced fluctuations throughout the acquisition, reflecting the instability introduced by repeatedly concentrating measurements in a restricted region, i.e. the model is well-calibrated locally but poorly constrained elsewhere. MU shows a smoother decrease, while BEACON achieves a comparably low MAE with less variance, suggesting that its spatially distributed sampling leads to a more globally accurate surrogate.

The surrogate mean (**Figure 3g**) and surrogate uncertainty (**Figure 3h**) together characterize the internal state of the model as learning progresses. For BEACON, both quantities stabilize relatively early and evolve smoothly, reflecting a model that is being updated with diverse, informative measurements. EI, by contrast, shows persistent large excursions in both mean and uncertainty, particularly visible as sharp spikes in **Figure 3h**. These spikes correspond to steps where the model encounters measurements that are poorly represented in the current training set, which is a direct consequence of the spatially collapsed trajectory periodically revisiting a region that is outlying relative to the broader distribution. MU occupies an intermediate position, with moderate fluctuations that diminish as the budget is exhausted.

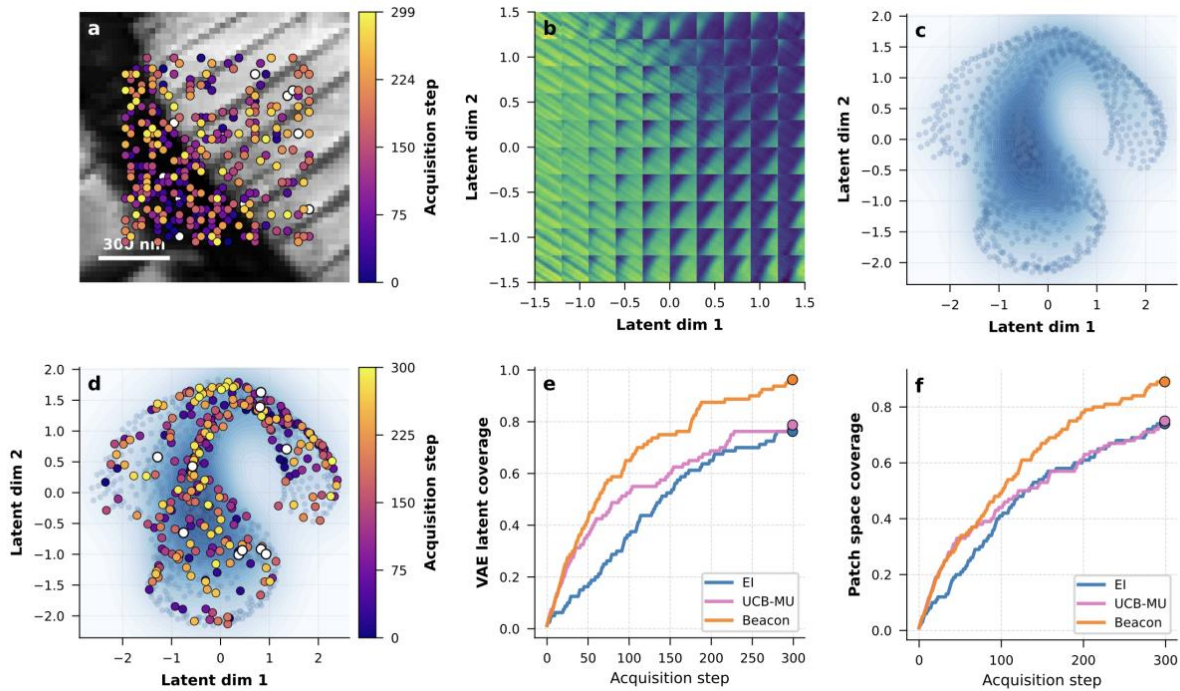

**Figure 4.** Exploration in real and VAE latent and patch space **(see definitions in supplementary).** (a) Beacon acquisition trajectory overlaid on the PFM image in real space, with marker color encoding acquisition step order and white markers denoting seed points. (b) 2D VAE latent space representation (i.e., decoding VAE latent space on a grid of points). (c) Distribution of all the patches in the 2D VAE latent space. (d) Acquisition trajectory mapped into the latent space, illustrating how Beacon progressively explores the manifold over 300 steps. (e) VAE latent space coverage as a function of acquisition step, quantifying the fraction of the latent manifold visited by each strategy. (f) Patch space coverage, measuring diversity of acquired spectra in the high-dimensional patch space. In (e–f), Beacon (orange) achieves consistently broader coverage than EI (blue) and MU (pink) across the full acquisition budget.

Further insight into the behavior of the different exploration strategies emerges from analysis of the trajectory in the feature space of the system. To construct this representation, we use a variational autoencoder (VAE) that maps each image patch to a low-dimensional latent code. For ease of visualization, we use a two-dimensional latent space, although in practice the latent dimensionality can be optimized depending on the amount of available data and the reconstruction quality required. The essential role of the autoencoder is to take image patches, encode them into a reduced latent representation, and then decode that latent representation back into the original feature space. The model is trained by minimizing the combined reconstruction loss and KL loss, so that the latent variables capture the dominant factors of variation in the data while remaining smoothly organized.

The real-space exploration trajectory is shown in **Figure 4a** and represents the physical motion of the probe over the sample. To move beyond physical coordinates and instead examine what kinds of objects are being sampled, the full set of available patches is encoded by the autoencoder, and the resulting two-dimensional latent representation is shown in **Figure 4b**. Here, the points on a square grid in the latent plane are decoded back into the original feature space, thereby revealing the correspondence between the two latent coordinates and the real microstructural motifs. As illustrated in the figure, high values of both latent variables correspond to regions associated with in-plane *a* domains. Regions with high latent dimension 2 but low latent dimension 1 correspond to dense *a-c* domain patterns, whereas regions with high latent dimension 1 but low latent dimension 2 correspond to single domain walls. A key advantage of this latent representation is that the autoencoder tends to disentangle the latent representations, in the sense that the major factors of variation in the data become associated with smoothly varying latent coordinates. This discussion is necessarily general, but it illustrates the central principle: the latent variables provide a compact, continuous representation of the dominant structural variations present in the image patches.

In VAE analysis, each image patch becomes a single point in latent space with some uncertainty(note we omit the uncertainty) and the full distribution of these points is shown in **Figure 4c**. Notably, this latent distribution has a fairly complex structure, reflecting the fact that some microstructural elements are statistically well represented in the dataset, whereas others are much less common. This makes the latent space particularly useful for analyzing exploratory behavior, because the progression of the automated experiment can now be visualized directly as a trajectory through the latent manifold, i.e., as the sequence of structural motifs selected for measurement.

This trajectory is shown in **Figure 4 d**. While some sequential ordering of coverage can be discerned, the overall path follows a relatively specific route through the latent space, illustrating that the active-learning algorithm has effectively zoomed in on selected classes of microstructural elements associated with behaviors of interest. In direct analogy with the real-space analysis, one can then define and compare measures of latent-space coverage and patch-space coverage, shown in **Figures 4e** and **4f**, respectively. These metrics quantify how broadly the experiment traverses the space of structural motifs rather than merely how far the probe moves physically. In all cases, BEACON shows superior behavior compared with EI and MU, indicating a broader and more balanced traversal of the accessible feature manifold. This, in turn, is consistent with a more robust exploratory dynamic, in which the algorithm continues to sample diverse classes of objects rather than collapsing prematurely onto a narrow subset of the available microstructure.

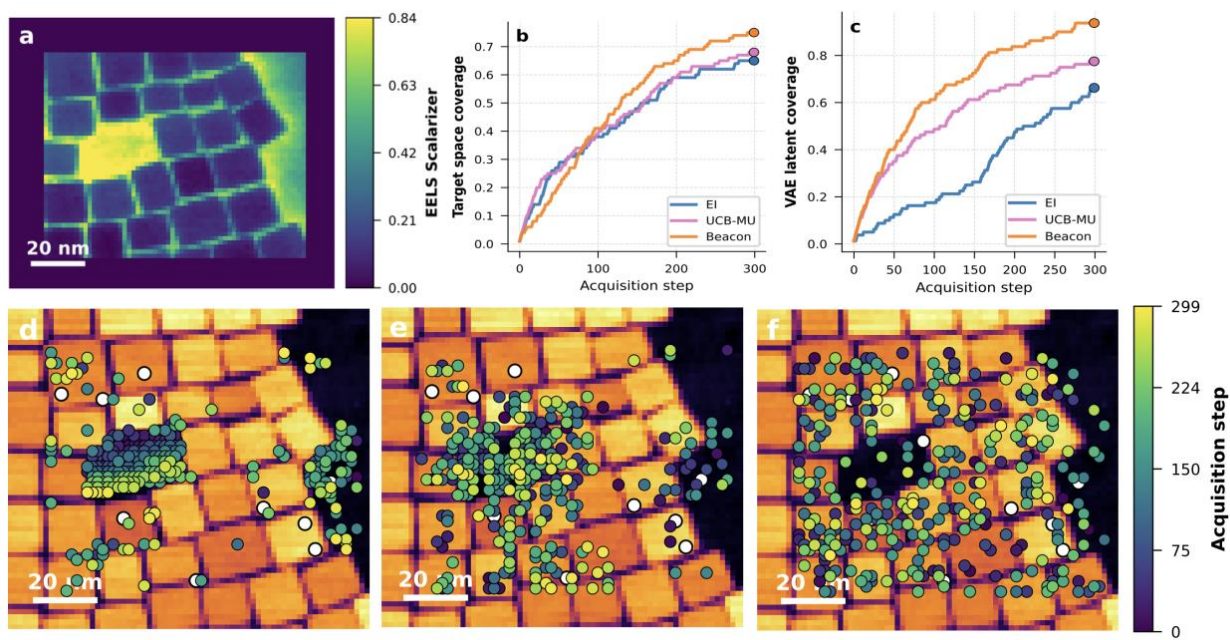

**Figure 5.** (a) Ground truth scalarizer map for the EELS data set, (b) Target space coverage as a function of acquisition step, showing the fraction of the target value distribution explored over time. (c) VAE latent space coverage as a function of acquisition step, quantifying the fraction of the latent manifold visited by each strategy. Show are exploratory trajectories in real space for (d) EI, (e) MU, and (f) BEACON acquisition functions. In (b-c) we see BEACON (orange) achieves consistently broader coverage than EI (blue) and MU (pink) across the full acquisition budget.

A similar pattern is observed for the STEM-EELS dataset, as illustrated in **Figure 5**. In this case, the image and the corresponding ground-truth behavior of the acquisition target are both comparatively simpler. The response is dominated by a relatively clear two-level contrast between the plasmonic nanoparticles and the regions between them, together with a more gradual decay of the plasmonic response along the outer edge of the dense particle assembly. This simpler structure provides a useful complementary case, since it allows the behavior of the different acquisition functions to be examined in a setting where the underlying target landscape is less intricate.

The evolution of the target-space coverage and the VAE latent-space coverage for BEACON, expected improvement, and MU is shown in **Figures 5b-c**. A notable feature of this dataset is that the target-space coverage appears broadly comparable across all three acquisition functions. In other words, when judged only by the range of scalarized responses that are discovered, the methods perform similarly. However, the distinction becomes much clearer in the feature-space coverage, where BEACON shows a clear advantage. This indicates that even when the algorithms reach similar levels of target-space discovery, they do so through very different modes of traversal of the underlying structural manifold.

This difference becomes even more apparent in the real-space trajectories shown in **Figures 5d-f**. For expected improvement, the algorithm clearly converges onto two localized regions of the sample surface. This is especially evident from the color gradient, where the transition from blue to yellow remains tightly confined to those same regions, indicating that once the algorithm identified these areas, it remained effectively locked there and continued to explore them systematically. The MU trajectory is more exploratory, but the sampling is still concentrated primarily in the central region, with only modestly broader coverage. In contrast, BEACON exhibits an almost uniform exploration of the available object space, with the trajectory distributed much more broadly across the sample.

As in the previous example, these real-space trajectories provide an independent measure of exploration quality. This is because the algorithm uses the structural content of the patch, but not the explicit spatial coordinates from which that patch was extracted. Therefore, any spatial concentration observed in the trajectory is an emergent property of the exploration dynamics rather than a direct consequence of the model input. In this sense, the broader spatial traversal produced by BEACON again indicates that it sustains more diverse exploration, even in a system where the target-space coverage alone might suggest only modest differences between acquisition strategies.

## IV. Implementation on STEM

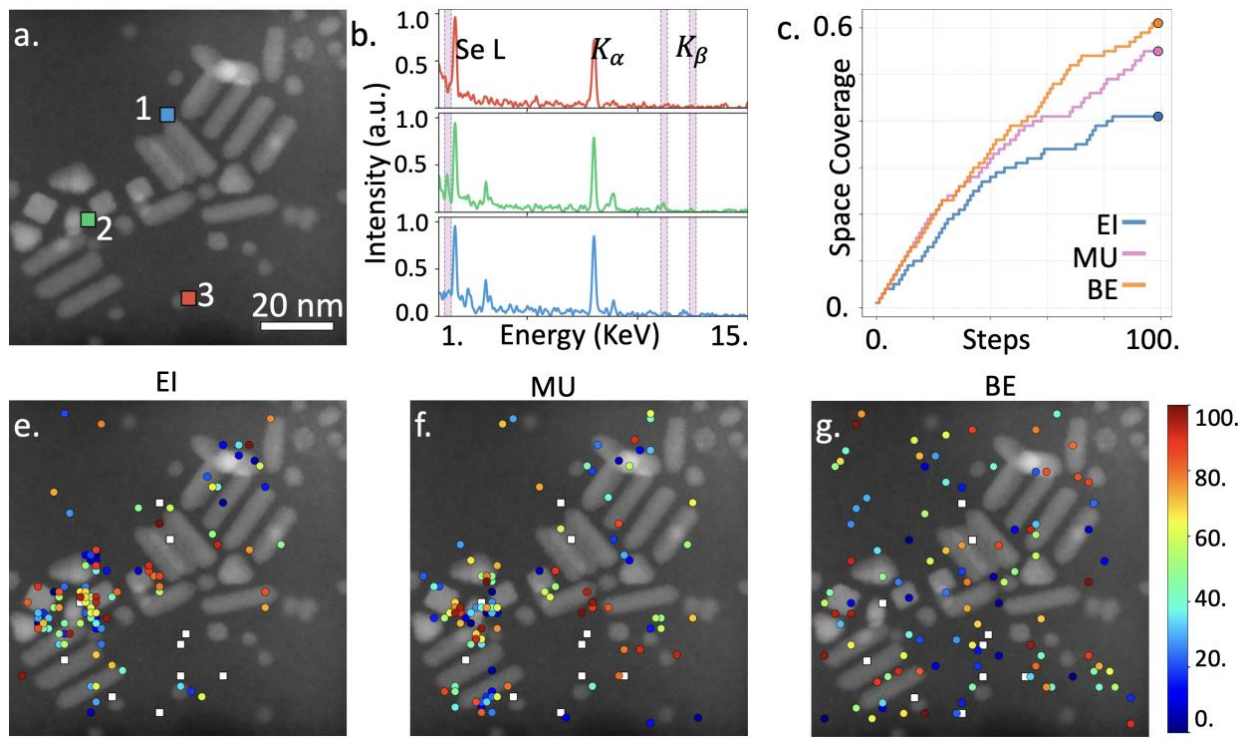

**Figure 5.** BEACON applied to live STEM-EDX acquisition on a Random library. (a) HAADF image with three representative measurement locations marked. (b) Normalized EDX spectra from the three locations, with Se L, $K_\alpha$, and $K_\beta$ emission lines indicated. (c) Patch space coverage as a function of acquisition step for EI, MU, and BEACON. (e–g) Acquisition trajectories for EI (e), MU (f), and BEACON (g) overlaid on the HAADF image, with marker color encoding step order and white squares denoting seed points.

To operationalize the novelty disco very in experimental settings, we deployed BEACON in a live autonomous STEM-EDX experiment on a mixed nanoparticle sample containing CdS nanorods (~6 × 18 nm and ~6 × 80 nm), CdSe nanoplatelets, and several quantum dot species (**Figure 5**). The scalarizer was defined from the integrated Se EDX signal, directing the active learning agent toward Se-rich nanostructures. Consistent with the pre-acquired dataset results, BEACON achieves substantially higher patch space coverage over 100 acquisition steps compared to both EI and MU (**Figure 5c**), confirming that the exploration advantage of the DKL-based diversity objective transfers directly to real experimental conditions. The acquisition trajectories (**Figure 5e-g**) further mirror the trends observed on PFM data: EI and MU trajectories show spatial clustering, while BEACON distributes measurements more uniformly across structurally distinct regions of the sample.

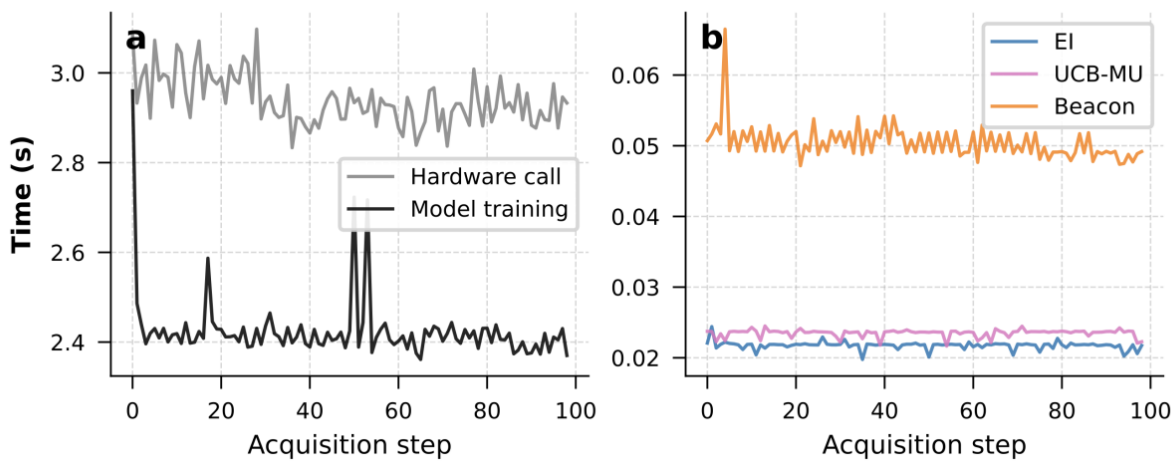

**Figure 6.** Computational timing per acquisition step. (a) Hardware and Model training time. (b) Acquisition function computation time for EI, MU, and Beacon.

All three strategies impose negligible computational overhead relative to the hardware acquisition time. As shown in **Figure 6,** the hardware step dominates at ~3 s per point, with Surrogate model training around ~2.4 s. Acquisition function evaluation is fast for all methods (Figure 5b), with EI and MU completing in ~0.02 s per step. Beacon requires marginally more compute (~0.05 s) due to Thompson sampling and computing the acquisition function. The timing will depend on the dataset size; currently, we use a 64*64 overview image.

**V. Summary**

Overall, we developed the deep-kernel-learning BEACON approach, implemented it on pre-acquired datasets, and provide the associated notebooks so that others can experiment with the workflow and deploy it on their own tools. We further operationalized it on the Spectra 300 microscope, illustrating how these ideas can transition from benchmarking on ground-truth datasets to execution in a real experimental setting. The results place BEACON in the broader context of a rapidly expanding ecosystem of active-learning and decision-making algorithms for automated experiments and, correspondingly, sharpen the need for clear, transparent ways to validate, compare, and benchmark their performance across representative datasets and use cases. Taken together, these developments point to an emerging "open era" of automated discovery, where algorithmic decisions become a first-class part of the measurement process rather than an afterthought applied only in post-processing.

Historically, exploration with imaging probes, whether in electron microscopy, scanning probe microscopy, nanoindenters, and related platforms, has been guided by a combination of prior human knowledge about which objects might matter and the distinctly human impulse to pursue unexpected behaviors. In practice, human decision-making is naturally anchored in what is immediately visible in the image space, because those features are available for direct inspection. Yet many discovery problems are defined not by what is apparent in the structural image, but by what emerges in the target space - often a spectral response that is only revealed sequentially and is not accessible to the human operator without substantial effort. Traditionally, the only reliable way to interrogate target-space behavior has been exhaustive mapping on a rectangular grid followed by post-acquisition analysis, often using unsupervised methods after the fact.

Deep kernel learning–type approaches change this operating mode by enabling the correlation between structure and properties to be learned on the fly, during the experiment, rather than retrospectively. Within this framing, BEACON provides a concrete implementation that moves beyond purely optimization-driven policies and instead supports systematic discovery of distinct behaviors in the target space. Rather than collapsing quickly into a narrow subset of the sample once a locally favorable region is identified, BEACON is designed to continue sampling in a way that promotes the identification of diverse target behaviors i.e., to find representative examples of the behavioral regimes present in the system.

At the same time, the results emphasize that "discovery" is not a single universal objective: it requires explicit choices about what constitutes novelty (in feature space, target space, or both), how discovery criteria are encoded, and which monitoring metrics are used to evaluate progress. In this study, the uniformity and mixing of exploration trajectories in image space served as a particularly useful independent indicator that BEACON sustained exploration rather than becoming trapped, even though such trajectory-based measures could themselves be elevated to optimization targets depending on the intended experimental goal.

**Data and Code availability:**

All the data and code are available at below GitHub repository:

https://github.com/utkarshp1161/Active-learning-in-microscopy/tree/beacon-dkl/notebooks/beacon-dkl

**Acknowledgements**

This work (BEACON DKL development) was supported (K.B., U.P., B.S., S.V.K., C.B., C.L., B.C.) by the U.S. Department of Energy, Office of Science, Office of Basic Energy Sciences as part of the Energy Frontier Research Centers program: CSSAS-The Center for the Science of Synthesis Across Scales under award number DE-SC0019288. The work was partially supported (U.P.) by AI Tennessee Initiative at University of Tennessee Knoxville (UTK). We gratefully acknowledge Kevin Roccapriore (ORNL) and Yongtao Liu (ORNL) for making their data publicly available for the benchmarking analysis.

# References

(1) Pennycook, S. J. , N. P. D. *Scanning Transmission Electron Microscopy*; Pennycook, S. J., Nellist, P. D., Eds.; Springer New York: New York, NY, 2011. https://doi.org/10.1007/978-1-4419-7200-2.

(2) Williams, D. B.; Carter, C. B. The Transmission Electron Microscope. In *Transmission Electron Microscopy*; Springer US: Boston, MA, 1996; pp 3–17. https://doi.org/10.1007/978-1-4757-2519-3_1.

(3) Crewe, A. V. Scanning Transmission Electron Microscopy*. *J. Microsc.* **1974**, *100* (3), 247–259. https://doi.org/10.1111/j.1365-2818.1974.tb03937.x.

(4) Krivanek, O. L.; Chisholm, M. F.; Nicolosi, V.; Pennycook, T. J.; Corbin, G. J.; Dellby, N.; Murfitt, M. F.; Own, C. S.; Szilagyi, Z. S.; Oxley, M. P.; Pantelides, S. T.; Pennycook, S. J. Atom-by-Atom Structural and Chemical Analysis by Annular Dark-Field Electron Microscopy. *Nature* **2010**, *464* (7288), 571–574. https://doi.org/10.1038/nature08879.

(5) Muller, D. A.; Kourkoutis, L. F.; Murfitt, M.; Song, J. H.; Hwang, H. Y.; Silcox, J.; Dellby, N.; Krivanek, O. L. Atomic-Scale Chemical Imaging of Composition and Bonding by Aberration-Corrected Microscopy. *Science (1979).* **2008**, *319* (5866), 1073–1076. https://doi.org/10.1126/science.1148820.

(6) Egerton, R. F. *Electron Energy-Loss Spectroscopy in the Electron Microscope*, Third edition.; Springer: New York, NY, 2011.

(7) Noircler, G.; Lebreton, F.; Drahi, E.; de Coux, P.; Warot-Fonrose, B. STEM-EELS Investigation of c-Si/a-AlO Interface for Solar Cell Applications. *Micron* **2021**, *145*, 103032. https://doi.org/10.1016/j.micron.2021.103032.

(8) Qu, J.; Sui, M.; Li, R. Recent Advances in In-Situ Transmission Electron Microscopy Techniques for Heterogeneous Catalysis. *iScience* **2023**, *26* (7), 107072. https://doi.org/10.1016/j.isci.2023.107072.

(9) Yu, L.; Li, M.; Wen, J.; Amine, K.; Lu, J. (S)TEM-EELS as an Advanced Characterization Technique for Lithium-Ion Batteries. *Mater. Chem. Front.* **2021**, *5* (14), 5186–5193. https://doi.org/10.1039/D1QM00275A.

(10) Nakamae, K. Electron Microscopy in Semiconductor Inspection. *Meas. Sci. Technol.* **2021**, *32* (5), 052003. https://doi.org/10.1088/1361-6501/abd96d.

(11) Schofield, S. R.; J Fisher, A.; Ginossar, E.; Lyding, J. W.; Silver, R.; Fei, F.; Namboodiri, P.; Wyrick, J.; Masteghin, M. G.; Cox, D. C.; Murdin, B. N.; Clowes, S. K.; G Keizer, J.; Y Simmons, M.; Stemp, H. G.; Morello, A.; Voisin, B.; Rogge, S.; A Wolkow, R.; Livadaru, L.; Pitters, J.; J Z Stock, T.; J Curson, N.; Butera, R. E.; V Pavlova, T.; Jakob, A. M.; Spemann, D.; Räcke, P.; Schmidt-Kaler, F.; Jamieson, D. N.; Pratiush, U.; Duscher,

G.; V Kalinin, S.; Kazazis, D.; Constantinou, P.; Aeppli, G.; Ekinci, Y.; Owen, J. H. G.; Fowler, E.; Moheimani, S. O. R.; Randall, J.; Misra, S.; A Ivie, J.; Allemang, C. R.; Anderson, E. M.; Bussmann, E.; Campbell, Q.; Gao, X.; Lu, T.-M.; Schmucker, S. W. Roadmap on Atomic-Scale Semiconductor Devices. *Nano Futures* **2025**, *9* (1), 012001. https://doi.org/10.1088/2399-1984/ada901.

(12) Dyck, O.; Ziatdinov, M.; Lingerfelt, D. B.; Unocic, R. R.; Hudak, B. M.; Lupini, A. R.; Jesse, S.; Kalinin, S. V. Author Correction: Atom-by-Atom Fabrication with Electron Beams. *Nat. Rev. Mater.* **2020**, *6* (7), 640–640. https://doi.org/10.1038/s41578-020-0188-y.

(13) Boebinger, M. G.; Brea, C.; Ding, L.; Misra, S.; Olunloyo, O.; Yu, Y.; Xiao, K.; Lupini, A. R.; Ding, F.; Hu, G.; Ganesh, P.; Jesse, S.; Unocic, R. R. The Atomic Drill Bit: Precision Controlled Atomic Fabrication of 2D Materials. *Advanced Materials* **2023**, *35* (14). https://doi.org/10.1002/adma.202210116.

(14) Zhou, M.; Zhang, W.; Sun, J.; Chu, F.; Dong, G.; Nie, M.; Xu, T.; Sun, L. Atomic Fabrication of 2D Materials Using Electron Beams Inside an Electron Microscope. *Nanomaterials* **2024**, *14* (21), 1718. https://doi.org/10.3390/nano14211718.

(15) Batson, P. E.; Dellby, N.; Krivanek, O. L. Sub-Ångstrom Resolution Using Aberration Corrected Electron Optics. *Nature* **2002**, *418* (6898), 617–620. https://doi.org/10.1038/nature00972.

(16) Gui, C.; Zhang, Z.; Li, Z.; Luo, C.; Xia, J.; Wu, X.; Chu, J. Deep Learning Analysis on Transmission Electron Microscope Imaging of Atomic Defects in Two-Dimensional Materials. *iScience* **2023**, *26* (10), 107982. https://doi.org/10.1016/j.isci.2023.107982.

(17) Yang, Z.; Yin, L.; Lee, J.; Ren, W.; Cheng, H.; Ye, H.; Pantelides, S. T.; Pennycook, S. J.; Chisholm, M. F. Direct Observation of Atomic Dynamics and Silicon Doping at a Topological Defect in Graphene. *Angewandte Chemie International Edition* **2014**, *53* (34), 8908–8912. https://doi.org/10.1002/anie.201403382.

(18) Roccapriore, K. M.; Torsi, R.; Robinson, J.; Kalinin, S.; Ziatdinov, M. Dynamic STEM-EELS for Single-Atom and Defect Measurement during Electron Beam Transformations. *Sci. Adv.* **2024**, *10* (29). https://doi.org/10.1126/sciadv.adn5899.

(19) Shen, M.; Li, G.; Wu, D.; Yaguchi, Y.; Haley, J. C.; Field, K. G.; Morgan, D. A Deep Learning Based Automatic Defect Analysis Framework for In-Situ TEM Ion Irradiations. *Comput. Mater. Sci.* **2021**, *197*, 110560. https://doi.org/10.1016/j.commatsci.2021.110560.

(20) Groschner, C. K.; Choi, C.; Scott, M. C. Machine Learning Pipeline for Segmentation and Defect Identification from High-Resolution Transmission Electron Microscopy Data. *Microscopy and Microanalysis* **2021**, *27* (3), 549–556. https://doi.org/10.1017/S1431927621000386.

(21) Volpe, G.; Wählby, C.; Tian, L.; Hecht, M.; Yakimovich, A.; Monakhova, K.; Waller, L.; Sbalzarini, I. F.; Metzler, C. A.; Xie, M.; Zhang, K.; Lenton, I. C. D.; Rubinsztein-Dunlop, H.; Brunner, D.; Bai, B.; Ozcan, A.; Midtvedt, D.; Wang, H.; Sladoje, N.; Lindblad, J.; Smith, J. T.; Ochoa, M.; Barroso, M.; Intes, X.; Qiu, T.; Yu, L.-Y.; You, S.; Liu, Y.; Ziatdinov, M. A.; Kalinin, S. V; Sheridan, A.; Manor, U.; Nehme, E.; Goldenberg, O.; Shechtman, Y.; Moberg, H. K.; Langhammer, C.; Špačková, B.; Helgadottir, S.; Midtvedt, B.; Argun, A.; Thalheim, T.; Cichos, F.; Bo, S.; Hubatsch, L.; Pineda, J.; Manzo, C.; Bachimanchi, H.; Selander, E.; Homs-Corbera, A.; Fränzl, M.; de Haan, K.; Rivenson, Y.; Korczak, Z.; Adiels, C. B.; Mijalkov, M.; Veréb, D.; Chang, Y.-W.; Pereira, J. B.; Matuszewski, D.; Kylberg, G.; Sintorn, I.-M.; Caicedo, J. C.; Cimini, B. A.; Bell, M. A. L.; Saraiva, B. M.; Jacquemet, G.; Henriques, R.; Ouyang, W.; Le, T.; Gómez-de-Mariscal, E.; Sage, D.; Muñoz-Barrutia, A.; Lindqvist, E. J.; Bergman, J. Roadmap on Deep Learning for Microscopy. 2023. https://arxiv.org/abs/2303.03793.

(22) Kalinin, S. V; Barakati, K.; Slautin, B. N.; Houston, A. C.; Pratiush, U.; Duscher, G.; Cao, C.; Sgouralis, I.; Spurgeon, S. R.; Yang, C.; Channing, G. Do We Really Need All That Data: From Data to Agency in Automated Microscopy. February 15, 2026. https://doi.org/10.26434/chemrxiv.15000070/v1.

(23) Alldritt, B.; Hapala, P.; Oinonen, N.; Urtev, F.; Krejci, O.; Federici Canova, F.; Kannala, J.; Schulz, F.; Liljeroth, P.; Foster, A. S. Automated Structure Discovery in Atomic Force Microscopy. *Sci. Adv.* **2020**, *6* (9). https://doi.org/10.1126/sciadv.aay6913.

(24) Ziatdinov, M.; Liu, Y.; Kalinin, S. V. Active Learning in Open Experimental Environments: Selecting the Right Information Channel(s) Based on Predictability in Deep Kernel Learning. 2022. https://arxiv.org/abs/2203.10181.

(25) Liu, Y.; Ziatdinov, M.; Kalinin, S. V. Exploring Causal Physical Mechanisms via Non-Gaussian Linear Models and Deep Kernel Learning: Applications for Ferroelectric Domain Structures. *ACS Nano* **2022**, *16* (1), 1250–1259. https://doi.org/10.1021/acsnano.1c09059.

(26) Thach, R. E.; Thach, S. S. Damage to Biological Samples Caused by the Electron Beam during Electron Microscopy. *Biophys. J.* **1971**, *11* (2), 204–210. https://doi.org/10.1016/S0006-3495(71)86208-2.

(27) Karuppasamy, M.; Karimi Nejadasl, F.; Vulovic, M.; Koster, A. J.; Ravelli, R. B. G. Radiation Damage in Single-Particle Cryo-Electron Microscopy: Effects of Dose and Dose Rate. *J. Synchrotron Radiat.* **2011**, *18* (3), 398–412. https://doi.org/10.1107/S090904951100820X.

(28) Liu, Y.; Pratiush, U.; Barakati, K.; Funakubo, H.; Lin, C.-C.; Kim, J.; Martin, L. W.; Kalinin, S. V. Domain Switching on the Pareto Front: Multi-Objective Deep Kernel

Learning in Automated Piezoresponse Force Microscopy. 2025. https://arxiv.org/abs/2506.08073.

(29) Liu, Y.; Kalinin, S. V. Pareto-Optimal Experimentation: Human-Guided Multi-Objective Bayesian Optimization in Scanning Probe Microscopy. *Nano Lett.* **2026**, *26* (1), 441–447. https://doi.org/10.1021/acs.nanolett.5c05373.

(30) Narasimha, G.; Kong, D.; Regmi, P.; Jin, R.; Gai, Z.; Vasudevan, R.; Ziatdinov, M. Multiscale Structure-Property Discovery via Active Learning in Scanning Tunneling Microscopy. 2024. https://arxiv.org/abs/2404.07074.

(31) Liu, Y.; Kelley, K. P.; Vasudevan, R. K.; Funakubo, H.; Ziatdinov, M. A.; Kalinin, S. V. Experimental Discovery of Structure–Property Relationships in Ferroelectric Materials via Active Learning. *Nat. Mach. Intell.* **2022**, *4* (4), 341–350. https://doi.org/10.1038/s42256-022-00460-0.

(32) Pratiush, U. Utkarshp1161/Active-Learning-in-Microscopy: V1.0.0. Zenodo December 2024. https://doi.org/10.5281/zenodo.14562670.

(33) Pratiush, U.; Austin Houston; Paolo Longo; Remco Geurts; Sergei Kalinin; Gerd Duscher. StemOrchestrator: Enabling Seamless Hardware Control and High-Throughput Workflows on Electron Microscopes. January 3, 2026. https://doi.org/10.31224/4645.

(34) Pratiush, U.; Duscher, G. StemOrchestrator. GitHub 2025. https://github.com/pycroscopy/pyAutoMic/tree/main/TEM/stemOrchestrator.

(35) Houston, A.; Pratiush, U.; Pycroscopy Developers. Asyncroscopy: Asynchronous Orchestration Framework for Automated Electron Microscopy. GitHub 2025. https://github.com/pycroscopy/asyncroscopy.

(36) Liu, Y.; Roccapriore, K.; Checa, M.; Valleti, S. M.; Yang, J.; Jesse, S.; Vasudevan, R. K. AEcroscopy: A Software–Hardware Framework Empowering Microscopy Toward Automated and Autonomous Experimentation. *Small Methods* **2024**. https://doi.org/10.1002/smtd.202301740.

(37) Pratiush, U.; Houston, A.; Kalinin, S. V.; Duscher, G. Realizing Smart Scanning Transmission Electron Microscopy Using High Performance Computing. *Review of Scientific Instruments* **2024**, *95* (10). https://doi.org/10.1063/5.0225401.

(38) Liu, Y.; Pratiush, U.; Bemis, J.; Proksch, R.; Emery, R.; Rack, P. D.; Liu, Y.-C.; Yang, J.-C.; Udovenko, S.; Trolier-McKinstry, S.; Kalinin, S. V. Integration of Scanning Probe Microscope with High-Performance Computing: Fixed-Policy and Reward-Driven Workflows Implementation. **2024**.

(39) Thermo Fisher Scientific. AutoScript TEM Software. 2025.

(40) de la Cruz, M. J.; Martynowycz, M. W.; Hattne, J.; Gonen, T. MicroED Data Collection with SerialEM. *Ultramicroscopy* **2019**, *201*, 77–80. https://doi.org/10.1016/j.ultramic.2019.03.009.

(41) Pycroscopy. PyAutoMic: DigitalMicrograph TEM Scripts. 2025.

(42) Slautin, B. N.; Pratiush, U.; Ivanov, I. N.; Liu, Y.; Pant, R.; Zhang, X.; Takeuchi, I.; Ziatdinov, M. A.; Kalinin, S. V. Co-Orchestration of Multiple Instruments to Uncover Structure–Property Relationships in Combinatorial Libraries. *Digital Discovery* **2024**. https://doi.org/10.1039/D4DD00109E.

(43) Pratiush, U.; Roccapriore, K. M.; Liu, Y.; Duscher, G.; Ziatdinov, M.; Kalinin, S. V. Building Workflows for Interactive Human in the Loop Automated Experiment (HAE) in STEM-EELS. **2024**.

(44) Liu, Y.; Yang, J.; Vasudevan, R. K.; Kelley, K. P.; Ziatdinov, M.; Kalinin, S. V.; Ahmadi, M. Exploring the Relationship of Microstructure and Conductivity in Metal Halide Perovskites via Active Learning-Driven Automated Scanning Probe Microscopy. *J. Phys. Chem. Lett.* **2023**, *14* (13), 3352–3359. https://doi.org/10.1021/acs.jpclett.3c00223.

(45) Ziatdinov, M.; Liu, Y.; Kelley, K.; Vasudevan, R.; Kalinin, S. V. Bayesian Active Learning for Scanning Probe Microscopy: From Gaussian Processes to Hypothesis Learning. *ACS Nano* **2022**, *16* (9), 13492–13512. https://doi.org/10.1021/acsnano.2c05303.

(46) Slautin, B. N.; Liu, Y.; Funakubo, H.; Kalinin, S. V. Unraveling the Impact of Initial Choices and In-Loop Interventions on Learning Dynamics in Autonomous Scanning Probe Microscopy. *J. Appl. Phys.* **2024**, *135* (15). https://doi.org/10.1063/5.0198316.

(47) Um, M.; Sanchez, S. L.; Song, H.; Lawrie, B. J.; Ahn, H.; Kalinin, S. V.; Liu, Y.; Choi, H.; Yang, J.; Ahmadi, M. Tailoring Molecular Space to Navigate Phase Complexity in Cs-Based Quasi-2D Perovskites via Gated-Gaussian-Driven High-Throughput Discovery. *Adv. Energy Mater.* **2025**, *15* (16). https://doi.org/10.1002/aenm.202404655.

(48) Biswas, A.; Vasudevan, R.; Pant, R.; Takeuchi, I.; Funakubo, H.; Liu, Y. SANE: Strategic Autonomous Non-Smooth Exploration for Multiple Optima Discovery in Multi-Modal and Non-Differentiable Black-Box Functions. *Digital Discovery* **2025**, *4* (3), 853–867. https://doi.org/10.1039/D4DD00299G.

(49) Pratiush, U.; Funakubo, H.; Vasudevan, R.; Kalinin, S. V.; Liu, Y. Scientific Exploration with Expert Knowledge (SEEK) in Autonomous Scanning Probe Microscopy with Active Learning. *Digital Discovery* **2025**, *4* (1), 252–263. https://doi.org/10.1039/D4DD00277F.

(50) Bulanadi, R.; Chowdhury, J.; Funakubo, H.; Ziatdinov, M.; Vasudevan, R.; Biswas, A.; Liu, Y. Beyond Optimization: Exploring Novelty Discovery in Autonomous Experiments.

*ACS Nanoscience Au* **2026**, *6* (1), 86–94. https://doi.org/10.1021/acsnanoscienceau.5c00106.

(51) Jolliffe, I. T.; Cadima, J. Principal Component Analysis: A Review and Recent Developments. *Philosophical Transactions of the Royal Society A: Mathematical, Physical and Engineering Sciences* **2016**, *374* (2065), 20150202. https://doi.org/10.1098/rsta.2015.0202.

(52) Kingma, D. P.; Welling, M. Auto-Encoding Variational Bayes. 2022. https://arxiv.org/abs/1312.6114.

(53) Slautin, B. N.; Pratiush, U.; Lupascu, D. C.; Ziatdinov, M. A.; Kalinin, S. V. Integrating Predictive and Generative Capabilities by Latent Space Design via the DKL-VAE Model. 2025. https://arxiv.org/abs/2503.02978.

(54) Rasmussen, C. E. Gaussian Processes in Machine Learning; 2004; pp 63–71. https://doi.org/10.1007/978-3-540-28650-9_4.

(55) Wilson, A. G.; Hu, Z.; Salakhutdinov, R.; Xing, E. P. Deep Kernel Learning. In *Proceedings of the 19th International Conference on Artificial Intelligence and Statistics*; Gretton, A., Robert, C. C., Eds.; Proceedings of Machine Learning Research; PMLR: Cadiz, Spain, 2016; Vol. 51, pp 370–378.

(56) Balandat, M.; Karrer, B.; Jiang, D. R.; Daulton, S.; Letham, B.; Wilson, A. G.; Bakshy, E. BoTorch: Programmable Bayesian Optimization in PyTorch. *ArXiv* **2019**, *abs/1910.06403*.

(57) Kalinin, S. V; Liu, Y.; Biswas, A.; Duscher, G.; Pratiush, U.; Roccapriore, K.; Ziatdinov, M.; Vasudevan, R. Human-in-the-Loop: The Future of Machine Learning in Automated Electron Microscopy. *Micros. Today* **2024**, *32* (1), 35–41. https://doi.org/10.1093/mictod/qaad096.

(58) Wu, X.; Xiao, L.; Sun, Y.; Zhang, J.; Ma, T.; He, L. A Survey of Human-in-the-Loop for Machine Learning. *Future Generation Computer Systems* **2022**, *135*, 364–381. https://doi.org/10.1016/j.future.2022.05.014.

(59) Pratiush, U.; Duscher, G.; Kalinin, S. Human-in-the-Loop Interface for Automated Experiments in Electron Microscopy, Automated Characterization. In *AI for Accelerated Materials Design - NeurIPS 2024*; 2024.

(60) Valleti, M.; Vasudevan, R. K.; Ziatdinov, M. A.; Kalinin, S. V. Deep Kernel Methods Learn Better: From Cards to Process Optimization. *Mach. Learn. Sci. Technol.* **2024**, *5* (1), 015012. https://doi.org/10.1088/2632-2153/ad1a4f.

(61) Pratiush, U.; Houston, A.; Liu, R.; Duscher, G.; Kalinin, S. Towards Self-Optimizing Electron Microscope: Robust Tuning of Aberration Coefficients via Physics-Aware Multi-Objective Bayesian Optimization. 2026. https://arxiv.org/abs/2601.18972.

(62) Liu, Y.; Ziatdinov, M. A.; Vasudevan, R. K.; Kalinin, S. V. Explainability and Human Intervention in Autonomous Scanning Probe Microscopy. *Patterns* **2023**, *4* (11), 100858. https://doi.org/10.1016/j.patter.2023.100858.

(63) Tang, W.-T.; Chakrabarty, A.; Paulson, J. A. BEACON: A Bayesian Optimization Strategy for Novelty Search in Expensive Black-Box Systems. 2025. https://arxiv.org/abs/2406.03616.

(64) Tang, W.-T.; Chakrabarty, A.; Paulson, J. A. TR-BEACON: Shedding Light on Efficient Behavior Discovery in High-Dimensional Spaces with Bayesian Novelty Search over Trust Regions. In *Proceedings of the NeurIPS 2024 Workshop on Bayesian Decision-making and Uncertainty*; 2024.

(65) Terayama, K.; Sumita, M.; Tamura, R.; Payne, D. T.; Chahal, M. K.; Ishihara, S.; Tsuda, K. Pushing Property Limits in Materials Discovery *via* Boundless Objective-Free Exploration. *Chem. Sci.* **2020**, *11* (23), 5959–5968. https://doi.org/10.1039/D0SC00982B.

(66) Gardner, J. R.; Pleiss, G.; Bindel, D.; Weinberger, K. Q.; Wilson, A. G. GPyTorch: Blackbox Matrix-Matrix Gaussian Process Inference with GPU Acceleration. In *Proceedings of the 32nd International Conference on Neural Information Processing Systems*; NIPS'18; Curran Associates Inc.: Red Hook, NY, USA, 2018; pp 7587–7597.

(67) Pycroscopy Developers. DTMicroscope: Digital Twin Microscope. GitHub 2025. https://github.com/pycroscopy/DTMicroscope.

(68) Liu, Y.; Kelley, K. P.; Vasudevan, R. K.; Zhu, W.; Hayden, J.; Maria, J.; Funakubo, H.; Ziatdinov, M. A.; Trolier-McKinstry, S.; Kalinin, S. V. Automated Experiments of Local Non-Linear Behavior in Ferroelectric Materials. *Small* **2022**, *18* (48). https://doi.org/10.1002/smll.202204130.

(69) Roccapriore, K. M.; Ziatdinov, M.; Cho, S. H.; Hachtel, J. A.; Kalinin, S. V. Predictability of Localized Plasmonic Responses in Nanoparticle Assemblies. *Small* **2021**, *17* (21). https://doi.org/10.1002/smll.202100181.

## Supplementary

To quantify behavior of active learning trajectory, we introduce several complementary metrics that characterize how the surrogate model evolves (discussed in S1) during the experiment and how they explore (discussed in S2.) the patch, feature and target space.

Implementation examples using pre-acquired data are provided as Jupyter notebooks in the accompanying code repository.

**S1. Definition of learning curves:**

Purpose of these is to monitor and diagnose the surrogate model during active learning. Has been also discussed in in this work[43].

S1. a. Mean absolute error of surrogate v/s steps

Mean absolute error (MAE) of surrogate, which evaluates how accurately the surrogate model reproduces the ground-truth scalarizer map at a given active-learning step. If $\mu_t(x_j)$ is the surrogate prediction at location $x_j$ after $t$ measurements, and $y(x_j)$ is the corresponding ground-truth scalarizer value, then over a benchmark dataset containing $N$ candidate locations one may define

$$\text{MAE}_t = \frac{1}{N} \sum_{j=1}^{N} |\mu_t(x_j) - y(x_j)|.$$

Lower values of $\text{MAE}_t$ indicate a more accurate global surrogate. This metric captures a different aspect of performance than target-space coverage: an algorithm may reduce MAE rapidly by focusing on a narrow but information-rich region yet still fail to explore the full diversity of behaviors present in the sample. Conversely, a method that prioritizes discovery may maintain broader target coverage while reducing MAE more gradually, especially in early stages. Thus, MAE quantifies the fidelity of the learned structure–property model, while coverage quantifies the diversity of discovered behaviors.

S1. b. Surrogate mean v/s steps

The surrogate predictive mean reflects the overall level of the property being mapped as estimated by the model at a given active-learning step. Averaging the pointwise predictions across all $N$ candidate locations gives a single scalar summary,

$$\bar{\mu}_t = \frac{1}{N} \sum_{j=1}^{N} \mu_t(x_j).$$

A rising $\bar{\mu}_t$ over steps indicates that the acquisition function is preferentially directing measurements toward high-value regions, consistent with exploitation-dominated behavior. A flat or slowly evolving $\bar{\mu}_t$, by contrast, suggests that the algorithm is sampling broadly across the property range rather than converging on a specific target. This metric is most informative when read alongside surrogate uncertainty $U_t$ and target-space coverage: rapid growth in $\bar{\mu}_t$ paired with declining coverage is a signature of premature exploitation, whereas a stable $\bar{\mu}_t$ combined with expanding coverage reflects healthy exploratory behavior.

S1. c. Surrogate uncertainty v/s steps

Surrogate predictive uncertainty, which reflects how uncertain the model remains across the candidate measurement space. If $\sigma_t(x_j)$ denotes the predictive standard deviation of the surrogate at location $x_j$, then a natural global summary is

$$U_t = \frac{1}{N} \sum_{j=1}^{N} \sigma_t(x_j).$$

A decreasing $U_t$ indicates that the model is becoming more confident as measurements accumulate. However, the interpretation of this trend requires care. A rapid drop in uncertainty can be beneficial if it reflects genuine learning over the full feature space, but it can also signal premature overconfidence caused by repeated sampling of a restricted region. Conversely, persistently high uncertainty may indicate that the algorithm continues to explore broadly, but it may also imply that the surrogate has not yet consolidated a stable global model. For this reason, uncertainty must be interpreted jointly with target-space coverage (**S2 c.**) and MAE: the most desirable behavior is not simply minimal uncertainty, but a balanced regime in which uncertainty decreases while the algorithm continues to expand coverage and improve predictive accuracy.

## S2. Definition of coverage:

The purpose of these metrics is to quantify the novelty and diversity of the active learning trajectory across three complementary spaces: the raw image patch space, the learned feature space, and the target (scalarizer) space. Together they provide a multi-scale picture of exploration that no single metric can capture alone.

We define patch space as the space of raw local image windows extracted from the scan, feature space as the low-dimensional latent representation learned by a variational autoencoder (VAE) trained on those patches, and target space as the scalar property values assigned to each location by the scalarizer function. Coverage in each space is computed as the fraction of discrete regions defined by $k$-means clusters or histogram bins, that have been visited by at least one acquired point up to step $t$.

S2 a. Patch space coverage:

Patch space coverage measures the structural diversity of the image windows selected by the active learning algorithm, without any learned representation. Each patch $p_j \in \mathbb{R}^d$ (where $d = H \times W$ for a window of height $H$ and width $W$) is first mean- and variance-normalized, then projected to a lower-dimensional space via a random linear projection, and finally assigned to one of $K$ clusters obtained by $k$-means on all $N$ patches. Letting $\ell(j) \in \{1, \dots, K\}$ denote the cluster label of patch $j$, patch-space coverage at step $t$ is defined as

$$C_t^{\text{patch}} = \frac{|\{\ell(j) : j \in \mathcal{A}_t\}|}{K},$$

where $\mathcal{A}_t$ denotes the set of acquired indices up to step $t$. A value of $C_t^{\text{patch}} = 1$, indicates that at least one patch from every structural cluster has been measured. This metric can be evaluated on both pre-acquired datasets and in real-time during live instrument operation, since it requires no ground-truth property labels.

S2 b. Feature space (VAE latent space) coverage:

Feature space coverage measures diversity in the learned latent representation of image patches. A two-dimensional variational autoencoder specifically, the invariant VAE (iVAE) implementation in pyroVED(https://github.com/ziatdinovmax/pyroVED), is trained on all $N$ patches simultaneously. Each patch is encoded to a latent mean vector $z_j \in \mathbb{R}^2$, and all latent vectors are clustered into $K$ groups using $k$-means. Feature-space coverage at step $t$ is then

$$C_t^{\text{feat}} = \frac{|\{\ell^z(j): j \in \mathcal{A}_t\}|}{K},$$

where $\ell^z(j)$ is the cluster label of patch $j$ in latent space. Compared to patch-space coverage, this metric is sensitive to semantically meaningful structural variation captured by the VAE rather than pixel-level differences, making it more robust to noise and imaging artifacts. Like patch-space coverage, it can be evaluated on pre-acquired datasets and during live experiments, as it depends only on the image data and not on measured property values.

S2 c. Target space coverage – Only on pre-acquired dataset

Target space coverage quantifies how much of the observable property range has been sampled by the active learning trajectory. The ground-truth scalarizer map is discretized into $B$ equal-width bins spanning $[\min_j y(x_j), \max_j y(x_j)]$, and only bins containing at least one ground-truth point are considered reachable. Target-space coverage at step $t$ is

$$C_t^{\text{target}} = \frac{|\{b: \exists\, j \in \mathcal{A}_t \text{ s.t. } y(x_j) \in \text{bin } b\}|}{B^*},$$

where $B^*$ is the number of non-empty bins. Unlike patch and feature space coverage, this metric requires access to the full ground-truth scalarizer map and is therefore only applicable to pre-acquired benchmark datasets rather than live instrument experiments.

Taken together, these six metrics provide a comprehensive and multi-scale diagnostic framework for evaluating active learning trajectories in scanning microscopy. The learning curve metrics — MAE, surrogate mean, and surrogate uncertainty, characterize the internal state of the probabilistic model and answer the question: is the surrogate learning effectively? The coverage metrics, in patch, feature, and target space, characterize the external behavior of the acquisition strategy and answer the complementary question: is the algorithm exploring meaningfully? Crucially, no single metric is sufficient on its own. An algorithm may achieve low MAE by repeatedly sampling a narrow but information-rich region while failing to discover the full diversity of structural behaviors present in the sample. Conversely, broad coverage in patch or feature space does not guarantee accurate property prediction if the acquired points are not informative for the surrogate. The most desirable active learning behavior is one in which MAE decreases steadily, uncertainty reduces without premature collapse, and coverage expands consistently across all three spaces simultaneously. By reporting all six metrics in parallel, one can distinguish between genuinely exploratory strategies and those that merely appear diverse in one space while remaining confined in another, a distinction that is especially consequential when deploying active learning on a physical instrument where each measurement carries a real experimental cost.